\documentclass[conference]{IEEEtran}
\IEEEoverridecommandlockouts
\usepackage{cite}
\usepackage{amsmath,amssymb,amsfonts}
\usepackage{algorithmic}
\usepackage{graphicx}
\usepackage{textcomp}
\usepackage{xcolor}
\def\BibTeX{{\rm B\kern-.05em{\sc i\kern-.025em b}\kern-.08em
    T\kern-.1667em\lower.7ex\hbox{E}\kern-.125emX}}
\usepackage{multirow}
\usepackage{blindtext, graphicx}
\usepackage{url}
\usepackage{tablefootnote}
\usepackage{listings}
\title{Benchmarks and Metrics for Evaluations of Code Generation: A Critical Review}

\author{
\IEEEauthorblockN{Debalina Ghosh Paul, Hong Zhu and Ian Bayley}
\IEEEauthorblockA{School of Engineering, Computing and Mathematics, Oxford Brookes University\\ Oxford OX33 1HX, UK. Email: hzhu@brookes.ac.uk}
}

\begin{document}

\maketitle
%\thispagestyle{empty}
%\pagestyle{empty}

%%%%%%%%%%%%%%%%%%%%%%%%%%%%%%%%%%%%%%%%%%%%%%%%%%%%%%%%%%%%%%%%%%%%%%%%%%%%%%
\begin{abstract}
With the rapid development of Large Language Models (LLMs), a large number of machine learning models have been developed to assist programming tasks including the generation of program code from natural language input. However, how to evaluate such LLMs for this task is still an open problem despite of the great amount of research efforts that have been made and reported to evaluate and compare them. This paper provides a critical review of the existing work on the testing and evaluation of these tools with a focus on two key aspects: the benchmarks and the metrics used in the evaluations. Based on the review, further research directions are discussed. 
\end{abstract}

\begin{IEEEkeywords} Machine learning, Large language models, Code generation, Performance evaluation, Benchmarks, Metrics.
\end{IEEEkeywords}
%%%%%%%%%%%%%%%%%%%%%%%%%%%%%%%%%%%%%%%%%%%%%%%%%%%%%%%%%%%%%%%%%%%%%%%%%%%%%%%%
\section{Introduction}

The recent years have seen a rapid growth in machine learning (ML) technology for natural language processing. Deep neural networks (DNN) in the transformer architecture \cite{g1} with billions of parameters have been developed. They have an impressive capability in conducting natural language processing (NLP) related tasks. Among the most valuable capabilities of these is code generation from natural language input. It is thought that this may fundamentally change the way software is developed; see e.g., \cite{g2}. A great amount of research efforts have been reported in the literature on uses of such general ML models and developing special purpose ones for solving programming problems; see, e.g. \cite{g3, g4, g5} for recent surveys of such ML models. 

However, it is still an open question how well they actually perform even after many reports have been published on this issue. A large number of different benchmarks and quality metrics have been proposed and employed in the evaluations. However, the conclusions of these evaluations and comparisons often conflict with each other, and the results hardly reflect the real experiences of the users. Therefore, it is highly desirable to understand the current state of the art in the evaluation of ML models as code generation tools. This paper provides such a review and analyses the directions for future research. 

The remainder of the paper is organised as follows. Section \ref{sec:LLMs} discusses various types of coding tasks that LLMs have been applied to solve, and summarises the large language models that are used or designed for solving coding problems. Section \ref{sec:Benchmarks} reviews the benchmarks used in the evaluations. Section \ref{sec:QualityAttributes} focuses on the quality attributes and their metrics of code generation. Section \ref{sec:Metrics} is devoted to the performance metrics. Section \ref{sec:Directions} analyses the problems in the current approach and discusses the directions for further research. 

\section{Overview of LLMs for Coding Tasks}\label{sec:LLMs}

The research on ML for generating program code can be backdated to 1980s; see, for example, \cite{r1, r2}. However, the emergence of LLMs has been a recent breakthrough. In this context, programming tasks are regarded roughly as translations between natural language descriptions of programming problems and codes in programming languages, or vice versa.  
%
%In this section, we will first identify and define various types of programming tasks where ML models have been employed to fulfil. We will then briefly reviews such ML models. Since the ML techniques underlying such models have been covered by existing surveys, we will only summarise their functionality to set the context for the review on the evaluation of them in the next sections. 
%
%Our focus will be on ML models after Dec. 2022 when ChatGPT 3 is announced, since earlier models are well covered by the existing surveys \cite{g3, g4, g5}. 
%
%\subsection{Programming Tasks} 
%
Therefore, such programming tasks can be classified into three categories: \emph{Description to Code} (D2C), \emph{Code to Description} (C2D) and \emph{Code to Code} (C2C) according to the source and target languages \cite{r4}. 

%\subsection{Programming Tasks} 

The D2C type of programming tasks take input in natural language that specifies the coding requirements; typically this is a functional specification. It is often called \emph{code generation} (CG) in the literature.  A typical example of the ML models for code generation is Codex\footnote{https://openai.com/index/openai-codex/} which underlies ChatGPT.  A special form of code generation is \emph{pseudo-code implementation}, which translates pseudo-code written in natural language text into program code \cite{c17}. It is worth noting that the input may also contain code fragments to define the context of the code to be generated. 

The C2D type of programming tasks take program code as the input and produce natural language text as the output. The typical examples of such tasks include \emph{document generation}, \emph{code summarisation} \cite{r6}, and \emph{comment generation} \cite{r5}, etc. 
%\begin{enumerate}
%\item \emph{Document Generation (DG)}, where the input is a piece of code and the ML model produces a document of the code to describe its functionality in natural language.  It is also called \emph{code summarisation} (CE) by many authors, since output from the ML model is essentially a summary of the code in natural language to explain how the code works \cite{r6}. 
%\item \emph{Comment Generation} (CmG), where the input is a piece of program code and the ML model adds comments in natural language to the code \cite{r5}. 
%\end{enumerate} 

The C2C type of programming tasks take a piece of code as the input and produce another piece of code as the output. Typical examples of this type include \emph{code completion} (such as GitHub's copilot\footnote{https://github.com/features/copilot}), \emph{code infilling} (like StarCoder\footnote{https://huggingface.co/bigcode/starcoder} \cite{r3}), \emph{code translation}, \emph{code refactoring}\cite{r8}, \emph{automatic debugging} (which is also called \emph{program repair} or \emph{code repair} by many authors) \cite{r7}, and \emph{test generation}\cite{r9,r10}, etc. 

In this paper, our focus is on the D2C type of programming tasks, but those issues that overlap with the C2C type will also be included; the C2D type will be omitted, however. 

%\subsection{Code Generation with General Purpose LLMs}

There are two approaches to developing ML models as tools for programming tasks. The first is to employ general purpose LLMs such as ChatGPT and Gemini \cite{c34}. Although their primary purpose lies in natural language processing, they also possess significant capability for various programming tasks since they have been trained on datasets that contain program code.

The second approach is to develop special purpose LLMs for programming tasks either through fine tuning of pre-trained LLMs or training the model from scratch.

Table \ref{tabIV} summarise the key features of the most well known LLMs for programming tasks, including their sizes, release years, the benchmarks used to evaluate their performance, and their performance as measured by $pass@100$ except for StarCoder, which was measured by $pass@1$. 

\begin{table}[h]
\caption{Performance Comparison of Language Models}
\label{tabIV}
\begin{center}
\begin{scriptsize}
\begin{tabular}{|l|l|c|c|l|r|}
\hline
\textbf{Model} & \textbf{Base Model} & \textbf{Size} & \textbf{Year} & \textbf{Benchmark} & \textbf{Score}\\
\hline
GPT-NEO\cite{c32}  & GPT-2 &  125M & 2021 & HumanEval  & 02.97 \\
                  &            &  1.3B &           &                      & 16.30 \\
                 &             &  2.7B &           &                      & 21.37\\
\hline
GPT-J\cite{c33} & GPT-2 &  6B & 2021 & HumanEval & 27.74\\
\hline
%Codex & GPT-3 &  12M & 2021 & HumanEval  & 08.58 \\
%          &             &  25M &          &                      & 12.89 \\
%           &             &  42M &          &                      & 15.55 \\
%          &             &  85M &          &                      & 22.40 \\
Codex\cite{c9} & GPT-3  &  300M &2021 & HumanEval  & 36.27\\
           &              &  679M &        &                    & 40.95\\
           &              &  2.5B  &         &                    & 59.50 \\
           &              &  12B   &         &                    & 72.31\\
\hline
TabNine & ? & ? & 2021 & HumanEval  & 7.59\\
\hline
ChatGPT & GPT-3.5 &  ? & 2022 & HumanEval  &   94.00\\
\hline
Gemini-Ultra\cite{c34} & Transformer  &  ? & 2023 & HumanEval  &  74.40\\
                                    &                       &    &          & Natural2Code  & 74.90\\
\hline
Gemini-Pro\cite{c34} & Transformer &  ? & 2023 & HumanEval  & 67.70 \\
                                  &                      &     &          & Natural2Code  & 69.60\\
\hline
CodeGen\cite{c35} & Auto-regressive &  350M & 2023 & HumanEval  &  35.19\\
                &  Transformer       &  2.7B &            &                      &  57.01\\
                &                           &   6.1B &            &                      &  65.82\\
                &                           &  16.1B &           &                      &  75.00\\
\hline
SantaCoder \cite{c36} & Decoder &  1.1B & 2023 & MultiPL-E  &   45.90\\
\hline
InCoder\cite{c37} &  Transformer  &  1.1B & 2023 & HumanEval  & 25.20 \\
              &                       &  6.7B &          &                      & 45.00\\
\hline
%StarCoder\cite{r3} & StarCoder-Base &  15.5B & 2023 & HumanEval   & 33.60\tablefootnote{\label{footnote1}Evaluated by $pass@1$} \\
%                 &            &            &          &   MBPP          & 52.70\footnotemark[\ref{footnote1}]\\
StarCoder\cite{r3} & StarCoder-Base &  15.5B & 2023 & HumanEval   & 33.60 \\
                 &            &            &          &   MBPP          & 52.70\\
\hline
\end{tabular}
\end{scriptsize}
\end{center}
\end{table}

While there is much research effort and literature on the evaluation of LLM performance, many research questions remain. For example, are the evaluations and comparisons fair and are the differences significant? Do the results of performance evaluation truly reflect the usability of  LLMs as practical programming tools? etc. In order to answer these questions, we will examine, in the subsequent sections, how the benchmarks were constructed, how the performances are assessed and the metrics are defined. 

\section{Benchmarks} \label{sec:Benchmarks}

Benchmarks play a crucial role in the evaluation of ML models and a large number of them have been proposed specifically for the evaluation of LLMs. Table \ref{tabI} summarises the most well known such benchmarks. We will discuss first how these benchmarks are constructed and then their main characteristics.  

%They are: APSS \cite{c8}, HumanEval \cite{c9}, MBPP \cite{c10}, MathQA-Python \cite{c10}, ClassEval \cite{c14}, CoderEval \cite{c5}, Multipl-E \cite{c11}, DS-1000 \cite{c7}, HumanEval+ \cite{c12}, CONCODE \cite{c6}, R-Benchmark \cite{c15}, JuICe \cite{c29}, Exec-CSN \cite{c30} and EvoCodeBench \cite{c31}.

To construct a benchmark, data must be procured, extracted and processed. We will consider each of these jobs in turn.

\subsection{Procurement}

We identify the following potential sources of data for programming tasks. Table \ref{tabI} also shows the sources of the benchmarks that we have reviewed. %Some of them have been used in the existing benchmarks, some have not. 

\begin{itemize}
\item \emph{Code repositories}, %which can be public open-source 
such as GitHub. 
%or private in-house owned by IT companies. A repository serves as a platform where developers can  access, contribute to, and collaborate on projects. It contains development tasks (or issues to be solved) and results (code for a coding task) of the completed tasks. The coding tasks in such a repository are mostly real programming problems. Thus, such data are valuable sources for code generation benchmarks. 
\item \emph{Online forums} for discussing programming problems and solutions, such as Stack Overflow. %An online question-answering forum serves as a platform that allow users to ask questions about programming languages, frameworks, tools, and other technical subjects, while other users provide solutions. Many questions are about how to code for a specific problem and the answers contain sample code. The origins of the questions can be from professional developers of different experiences as well as students and the answers could be of different qualities, too. 
\item \emph{Coding challenge sites}, %which are platforms that host coding competitions, 
such as Codewars, AtCoder, Kattis, and Codeforces. 
%The challenges are mostly in the form of coding problems and the participants of the challenge competitions provide solutions that contain program code. Due to the nature of competition, the coding tasks tend to be hard. It is questionable if such questions represents the real coding tasks in practice. 
\item \emph{Freelancer sites}, where software development tasks were outsourced, such as Upwork. %Most of the  tasks are from real development projects and the corresponding delivered results are codes of good quality. Such data are valuable as source of benchmark. However, some questions may originate from student coursework. Moreover, while the tasks are publicly available, the solutions are often not. 
\item \emph{Pre-existing datasets}, which can be included completely or partly in another benchmark. 
\item \emph{Textbooks} on programming. %, which contain programming exercises and many textbooks offer solutions to such exercise tasks.  The exercise questions are normally well presented and answers usually carefully tested and correct. However, these questions often focus on the usages of various language facilities and programming skills. They may be not representative to real development problems. Those textbooks targeting beginners tend to have very easy coding questions. 
\item \emph{Online Tutorial} websites, such as W3C resources. %that provide tutorial and educational material. Typically, they contain examples of code for various language facilities and their usages as well as various coding problems and the solutions. They also contains exercises questions and sample solutions. Such examples and exercises can be a source of benchmark data. 
\item \emph{Domain experts}, who custom-write tasks and provide solutions. % in the construction of a benchmark dataset. It is very expensive approach and hard to obtain large scale datasets. It is also hard to justify whether the datasets represents real problems. 
\item \emph{Crowdsourcing sites}, which gather data from the crowd. %, i.e. individuals or groups from the public. For example, the construction of MBPP \cite{c10} employed this approach with individuals from the crowd possessing programming knowledge. It is an economical method, but hard to control the quality of the data. 
\end{itemize}

%The sources for the benchmarks we have reviewed are summarised in Table \ref{tabI}. 

\begin{table}[h]
\caption{Sources of Existing Benchmarks}
\label{tabI}
\begin{center}
\begin{tabular}{|l|l|p{2.5 cm}|}
\hline
\textbf{Benchmark} & \textbf{Source} \\
\hline
APPS \cite{c8} & Coding challenge \\
\hline
HumanEval \cite{c9} & Domain Experts \\
\hline
MBPP \cite{c10} & Crowd-sourcing \\
\hline
MathQA-Python \cite{c10} & Dataset: MathQA \\
\hline
ClassEval \cite{c14} & Repository, Datasets: HumanEval, MBPP\\
\hline
CoderEval \cite{c5} & Repository: Github \\
\hline
Multipl-E \cite{c11} & Datasets: HumanEval, MBPP \\
\hline
DS-1000  \cite{c7}  & Forum: Stack Overflow \\
\hline
HumanEval+ \cite{c12} & Dataset: HumanEval  \\
\hline
CONCODE \cite{c6} & Repository: Github \\
\hline
R-benchmark \cite{c15} & Text Books \\
\hline
JuICe \cite{c29} &  Repository: Github (Jupyter, nbgrader)\\
\hline
Exec-CSN \cite{c30} & Repository: Github, Dataset: CodeSearchNet \\
\hline
EvoCodeBench \cite{c31} &  Repositories: GitHub\\
\hline
\end{tabular}
\end{center}
\end{table}

%Most sources required manual editing, as we will discuss but the use of parsers removed this need completely for Multipl-E and partially for APPS. In addition, the text book exercises used for R-benchmark were articulated so clearly that no editing was required once they were manually entered.

\subsection{Data Extraction and Processing}

Manual extraction from sources, as done for datasets CoderEval and DS-1000, is labour intensive. Automated extraction is less so and possible for online platforms like code repository and Q\&A forum, but work must still be done to write scripts or code for each source and to clean the data afterwards. Note that there is also the possibility of missing data.  When existing benchmarks are being reused, as with MathQA-Python, MultiPL-E and HumanEval+, extraction is easier to do. 

%\subsubsection{Data Processing}

Data extracted from a source often require processing before including in the benchmark. According to the purposes, data processing tasks can be classified into the following three types.

\begin{itemize}
\item \emph{Clarification}: clarifying the task description to reduce the ambiguity and incompleteness in the natural language specification of the task. Except for MultiPL-E, all benchmarks we reviewed have been manually edited after the data were extracted. For example, CoderEval employed 13 people for the task. When many people are involved, there is a risk of inconsistency between editors. 
\item \emph{Deduplication}: removing duplicated data from the dataset. It is done manually for DS-1000 and with automation for APPS, which employed tf-idf features coupled with SVD dimensionality reduction and cosine similarity. However, it is not clear whether and how this was done for the other benchmarks. 
\item \emph{Decontamination}: removing data used in the fine-tuning or training the LLM in the case that the benchmarks have been leaked to the LLM. This is done for APPS \cite{c8}. 
\end{itemize}

\subsection{Functionality and Structure}

One major difference between the benchmarks is the level of code generation: whether the task is to generate a statement, a function, a class or a whole program. This level of functionality is documented in Table \ref{tabII} below, along with the number of tasks in the benchmark and the programming language.

\begin{table}[h]
\caption{Functionality of Each Benchmark}
\label{tabII}
\begin{center}
\begin{tabular}{|l|l|c|l|}
\hline
\textbf{Benchmark} &  \textbf{Level}& \#\textbf{Tasks} &\textbf{Language}\\
\hline
APPS &Program  &10,000 &Python\\ \hline 
R-benchmark & Program &351 &R\\ \hline
ClassEval &Class &100 &Python\\ \hline
HumanEval &Function &164 &Python\\ \hline
MBPP &Function & 974 &Python\\ \hline
MathQA-Python &Function &23,914 &Python\\ \hline
CoderEval &Function & 230 &Python\\
                  &Method &230 &Java \\ \hline
Multipl-E &  Function & 1138 &Various\\ \hline
HumanEval+ &Function & 164 &Python\\ \hline
CONCODE &  Method & 2000 &Java\\ \hline
DS-1000 &Statement &1000 &Python\\ \hline
JuICe & Program & 3700 &Python\\ \hline
Exec-CSN & Function & 1931 & Python \\ \hline
EvoCodeBench & Function & 275 & Python \\ \hline
\end{tabular}
\end{center}
\end{table}

Another major difference between benchmarks is in the structure of the elements in the dataset. There are four types of components that have been included in existing benchmarks. Natural language descriptions are usually given as text. In addition, context code (such as function signatures) and unit test cases may be provided, both of which can be used for test automation to check the correctness of the generated solutions. In some cases, there may also be reference solutions. Table \ref{tabIII} gives the components provided for each benchmark, where the column \#Test Cases gives the average number of test cases per task if test cases are provided in the benchmark.

\begin{table}[h]
\caption{Structure of Data for Each Benchmark}
\label{tabIII}
\begin{center}
\begin{tabular}{|l|c|c|c|}
\hline
\textbf{Benchmark} &\textbf{Context Code} &\#\textbf{Test Cases} &\textbf{Solution}\\
\hline
APPS &-  &+ &+ %docstring, test cases, ground truth solutions   
\\ \hline
HumanEval &function signature  &7.7 &- %function signature, docstring, tests cases 
\\ \hline
MBPP &-  &3 &- %docstring, test cases 
\\ \hline
MathQA-Python &-  &3 &- %docstring, test cases 
\\ \hline
ClassEval &class skeleton  &33.1 &+ %class skeleton, method signatures, test cases, solution 
\\ \hline
CoderEval &function signature  &+ &- %function signature, docstring, tests cases 
\\ \hline
Multipl-E &function signature  &3 to 7 &- %function signature, docstring (in few cases), tests cases 
\\ \hline
DS-1000 &signature  &1.6 &+ %context, reference solution, test cases 
\\ \hline
HumanEval+ &function signature &774.8 &- %function signature, docstring, tests cases 
\\ \hline
CONCODE &function signature  &- &- %function signature, variables, docstring 
\\ \hline
R-benchmark &-  &- &+ %ID, Type, Difficulty, Source, input, output  
\\ \hline
JuICe &-  &- &+ %ID, Type, Difficulty, Source, input, output  
\\ \hline
Exec-CSN &function signature  &+ &+ %ID, Type, Difficulty, Source, input, output  
\\ \hline
EvoCodeBench &function signature  &+ &+ %ID, Type, Difficulty, Source, input, output  
\\ \hline
\end{tabular}
\end{center}
\end{table}

\subsection{Task Classification and Metadata}

In some cases, programming tasks in the dataset are classified into subsets of different difficulty levels. For example, Hendrycks et al. \cite{c8} distinguish three levels (Introductory, Interview, Competition) in APPS. Austin et al. \cite {c10} split tasks into two subsets MBPP and MathQA-Python as two levels of difficulty. Similarly, Yu et al. \cite{c5} used six levels according to the function's contextual dependency in CoderEval. 

Miah and Zhu \cite{c15} also distinguish five difficulty levels but they indicate the difficulty level as a part of the metadata associated to each task together with the task source and task types, making it possible to evaluate the effect of changing both of these. This can provide strong support to scenario-based testing and evaluation as shown in their case study. 

\section{Quality Attributes and Metrics}\label{sec:QualityAttributes}

We now review the quality attributes that LLMs are assessed against and the metrics used to measure LLMs. 
%The development of benchmarks for code generation also advanced the way code generators are evaluated. In general, the evaluation of a ML model involves two levels of activities. At the individual task level, the ML model under test is invoked on the test cases in a benchmark one by one and the quality of the output on each test case is assessed separately. At the benchmark level, the overall performance of the ML model is calculated from the assessments on individual test cases. In our case, the ML model is a code generator. Thus, the test cases are coding tasks and the outputs are program code, which could be accompanied by text explanations sometimes. Various techniques and methods have been developed to assess the quality of the output from an LLM on a coding task. 

\subsection{Functional Correctness}

Correctness of generated code is the main quality attribute for assessing the response of an LLM to a programming task, and according to what is provided by the benchmark, it is measured using reference solutions (ConCode), test cases (HumanEval, HumanEval+, MBPP, MathQA-Python and MultiPL-E) or a combination of both (APPS, ClassEval, CoderEval, and DS-1000). Where reference solutions are provided, correctness can either be functional correctness, as measured by passing tests, or syntactic closeness, for example, measured with the BLEU metric. 

Given a set $T_p = \{t_1, \cdots, t_n\}$ of test cases for a programming task $p$, a program code $P$ generated by a ML model $M$ is regarded as correct with respect to $T_p$, if $P$ is correct on all test cases $t_i$ in $T_p$. Let $B$ be a benchmark dataset for evaluating a ML model $M$. A basic performance metric based on pass-all-tests is the percentage of tasks in $B$ that the generated code successfully passes all tests. Note that there is randomness that an LLM generates program codes. This issue is addressed in the $pass@k$ metrics discussed in Section \ref{sec:Metrics}. 

Another metric commonly used is the percentage of test cases in $T_p$ on which the program $P$ is correct. This is denoted by $TPR_{T_p}(P)$, where $TPR$ stands for \emph{test pass rate}. The corresponding overall performance of the model $M$ on benchmark $B$ is the average test pass rate. Formally, 
\[AvgTPR_B(M)=\frac{\sum_{p\in B}{TPR_{T_p}(M(p))}}{\|B\|}\]

\subsection{Syntactic Closeness}

Similarity metrics have been used since the early evaluation of code generation. Some, like BLEU and ROGUE, were inherited from natural language, while others have been proposed specifically for code, such as Ruby and CodeBLEU. 

\subsubsection{BLEU} 

In 2002, Papineni et al. \cite{c16} proposed BLEU (Bilingual Evaluation Understudy) to evaluate the machine translations from one language to another. BLEU compares n-grams ($n$ number of contiguous words) between the generated translations by the language models with the reference translations. It calculates a precision score based on the number of n-grams of generated text $G$ that match with the reference text $R$, out of all generated n-grams. Then, this precision score is adjusted by a brevity penalty to account for translation length, penalising systems that generate excessively short translations. 

Formally, let $T$ be any given text that consists of a sequence of  words $\left(\tau_1, \cdots , \tau_k \right)$, where $k \geq 0$. The n-gram of $T$ can be defined as the set $Gram_n(T) = \{\left(\tau_i, \cdots, \tau_{i+n} \right) | i=1, \cdots, k-n\}$. The n-gram precision of text $G$ with respect to $R$ is defined as follows. 

\[p_n(T,R) = \frac{\|Gram_n(T) \cap Gram_n(R)\|}{\|Gram_n(T)\|}\]

The BLEU score is computed using the following formula. 
\[
BLEU(G,R) = BP \cdot \exp \left( \sum_{n=1}^{N} w_n \log p_n(G, R)\right)
\]
Here, \( w_n \) is the weight for n-gram precision, and \( BP \) is the brevity penalty,  which is computed using the following formula.
\[
BP = \begin{cases}
1 & \text{if } c > r \\
e^{(1 - \frac{r}{c})} & \text{if } c \leq r
\end{cases}
\]
where \( c \) is the length of the candidate translation and \( r \) is the length of the reference translation. 

The BLEU score is a number from 0 to 1 where a higher score indicates better alignment between the generated and reference translations.

\subsubsection{ROUGE} 

In 2004, Lin et al. \cite{c24} proposed ROUGE (Recall-Oriented Understudy for Gisting Evaluation) which is a family of metrics for comparing generated text with a set of reference texts.

ROUGE-N measures the overlap between the n-grams of a generated text $G$ and the n-grams of a set of  reference texts $R$, and is calculated using the following formula.
\[
ROUGE_N(G, R) = 
\frac{\sum_{\substack{S \in R}} \| Gram_n(S) \cap Gram_n(G) \|}{\sum_{\substack{S \in  R}} \| Gram_n(S) \| } 
\]

ROUGE-L measures the longest common subsequence (LCS) of words between the system generated and the reference texts. The calculation incorporates the precision \(P_{\text{lcs}}\) and recall \(R_{\text{lcs}}\) between the generated text \(G\) and reference text \(R\).  Let $len(T)$ be the length of text $T$, i.e. the number of words in $T$.  Formally, 
\[P_{\text{lcs}}(G, R) = \frac{LCS(G, R)}{len(G)}, ~~~R_{\text{lcs}}(G, R) = \frac{LCS(G, R)}{len(R)}\]
\[ROUGE_L(G, R) = \frac{(1 + \beta^2) \cdot P_{\text{lcs}}(G, R) \cdot R_{\text{lcs}}(G, R)}{R_{\text{lcs}}(G, R) + \beta^2 \cdot P_{\text{lcs}}(G, R)}
\]
\noindent{where $LCS(G, R)$ is the length of the longest common subsequence between $G$ and $R$, and $\beta$ is a parameter that balances precision and recall. Typically, $\beta = 1$ for equal weighting.}

There are three more metrics in the ROUGE family. ROUGE-W is a weighted variant of ROUGE-L, which assigns different weights to different LCS matches. ROUGE-S measures the overlap of skip-bigrams, which are any pair of words that occur in their sentence order, with allowance for arbitrary gaps. Skip-bigram co-occurrence statistics quantify the similarity in skip-bigrams between a generated and a set of reference texts. ROUGE-SU is a variant of ROUGE-S that also includes unigrams (single words). These metrics have not been used in the evaluation of code generation, so their definition is omitted. Readers are referred to \cite{c24} for details. 

\subsubsection{METEOR} 

In 2005, Banerjee et al. \cite{c25} proposed the METEOR (Metric for Evaluation of Translation with Explicit ORdering) metric, later extended by Denkowski et al. \cite{c26} to arbitrary  target languages. The generated and reference text are compared based on several components, including exact word matches, stemmed matches, semantic similarity using WordNet and phrase matches. Exact match counts the number of exact word matches between the two texts. Stemmed match captures variations of words that have the same root. Synonymy uses WordNet to find synonyms, and phrase match utilises a paraphrase table to find paraphrases. The METEOR score is computed using the harmonic mean of precision and recall (F-mean) and adjusted by a penalty for word order errors.
\[
METEOR = F_{\text{mean}} \cdot (1 - Penalty)
\]

The precision of a generated text $G$ with respect to a reference text $R$ is defined as the ratio of the number of  words in $G$ that matches words in the reference text $R$ (including exact, stemmed, and synonym matches) to the total number of words in $G$. Formally, let $MatchWords(G,R)$ denote the set of words in $G$ that matches words in $R$. and $\|T\|$ be the number of words in a text $T$. Formally, 
\[Prec(G,R) = \frac{\|MatchWords(G,R)\|}{\|G\|}\]

The recall is defined as the ratio of the number of words in $G$ that matches words in $R$ to the total number of words in the reference text $R$.
\[Rec(G,R) = \frac{\|MatchWords(G,R)\|}{\|R\|}\]

Penalty is applied for alignment fragmentation, which considers gaps and shifts in word order between the generated and reference texts. The penalty reduces the score for disorganized alignments.

\subsubsection{ChrF} 

In 2015, Popovic et al. \cite{c27} proposed the ChrF (Character n-gram F-score) metric which measures the similarity between a generated text and a reference text by comparing character n-grams (contiguous n character) instead of word-level tokens. The ChrF score is also calculated using the harmonic mean of precision and recall.

\[ChrF = \left(1 + \beta^2\right) \cdot \frac{Prec \cdot Rec}{\beta^2 \cdot Prec + Rec}\]
where:
\begin{itemize}
    \item \( Prec \) denotes precision, which is the ratio of the number of matched character n-grams to the total number of character n-grams in the generated text.
    \item \( Rec \) denotes recall, which is the ratio of the number of matched character n-grams to the total number of character n-grams in the reference text.
    \item \( \beta \) is a parameter that determines the relative importance of recall over precision (commonly set to 1 for balanced importance).
\end{itemize}

Ruby and CodeBLUE are custom metrics designed to address the characteristic features of programming languages. 

\subsubsection{Ruby} 

In 2019, Tran et al. \cite{c28} proposed Ruby, a similarity metric that compares the Program Dependency Graphs (PDGs) of the generated and reference codes. If a PDG cannot be constructed, then Abstract Syntax Trees (ASTs) are compared instead. If an AST cannot be constructed, the metric uses weighted string edit distance between the tokenized reference \(R\) and generated \(G\) codes. 
\[
RUBY(G, R) =
\begin{cases} 
GRS(G, R) & \text{if PDGs are applicable}, \\
TRS(G, R) & \text{if ASTs are applicable}, \\
STS(G, R) & \text{otherwise}.
\end{cases}
\]
where:
\begin{itemize}
  \item $GRS(G, R)$ measures the similarity between two program dependency graphs for $ R$  and  $G$.
  \item $TRS(R, C)$ measures the similarity between the Abstract Syntax Trees (ASTs) for $R$ and $G$.
  \item $STS(R, C)$ measures the string edit distance between $R$ and $G$.
\end{itemize}

\subsubsection{CodeBLEU} 

In 2020, Ren et al. \cite{c18} proposed CodeBLEU, a metric that extends the traditional BLEU metric by including program-specific features. CodeBLEU evaluates the similarity between the generated and reference codes by considering four different sub metrics: n-gram match (BLEU), weighted n-gram match, AST (Abstract Syntax Tree) match and data flow match. Weighted n-gram extends the traditional n-gram matching by assigning different weights to different types of tokens (e.g., keywords, identifiers, operators). AST match compares the Abstract Syntax Trees (ASTs) of the generated and reference codes. Data flow match evaluates the similarity in data flow graphs between the generated and reference code.
\begin{align*}
CodeBLEU = & \ \alpha \cdot BLEU + \beta \cdot \text{Weighted-N-gram} \\
& + \gamma \cdot \text{AST-Match} + \delta \cdot \text{DataFlowMatch}
\end{align*}
where \( \alpha \), \( \beta \), \( \gamma \) and \( \delta \) are weights that add up to 1.

In addition to the above metrics on the syntax closeness, Lai et al. used a much-relaxed form of similarity metric called surface-form constraints, which requires the presence or absence of certain specific APIs and/or the keywords in the solution code \cite{c7}. 

\subsubsection{Validity of the Metrics}

Although the use of similarity metrics in the performance evaluation of NLP models has been the standard approach in ML research, several authors have questioned if it is valid for code generation. In 2019, Kulal et al. \cite{c17} found that BLEU fails to assess functional correctness. In 2021, Hendrycks et al. \cite{c8} further showed it is inversely correlated with functional correctness. 

%More importantly, however, LLM evaluation metrics are only valid if the results of evaluation correctly reflect the perception of human users.

Evtikhiev et al. \cite{r11} is perhaps the first systematic study of the applicability of six similarity metrics, BLEU, ROUGE-L, METEOR, ChrF, CodeBLEU, and RUBY, to code generation. They investigated whether evaluations that employ these metrics yield statistically significant results and whether they correlate well with human judgement. Their conclusion was that an improvement of a corpus-level metric score by less than 2 points might not be enough to warrant a statistically significant improvement in quality without additional statistical tests. Even an improvement in score by less than 5 points may not correspond to a statistically significant improvement. They found that ChrF is the closest match to human assessment but it cannot be considered the ``perfect'' metric for code generation and such a metric is yet to be found. Interestingly, RUBY and CodeBLEU metrics, both developed for the specific purpose of  assessing code, performs no better than more generic metrics from the domain of machine translation. 

Kulal's solution to the problems observed with BLEU was to judge the correctness of an LLM generated solution by whether it passes all test cases. Hendrycks et al.  \cite{c8} also did this and additionally, used the test pass rate as a metric of correctness. Since then, most evaluations of code generation performances have employed test correctness. 

However, more recently, in their study of GitHub Copilot, Ziegler et al. found that ``while suggestion correctness is important, the driving factor for these improvements [in productivity] appears to be not correctness as such, but whether the suggestions are useful as a starting point for further development" \cite{r12}. This supports the assumption made in Miah and Zhu's study of ChatGPT's usability \cite{c15} that users may accept a generated solution if it is good enough to use even if it is not correct. 
%Table \ref{tabIV} summarises the numbers of these for each benchmark. 
%
%\begin{table}[h]
%\caption{Unit Tests and Reference Solutions for each Benchmark}
%\label{tabIV}
%\begin{center}
%\begin{tabular}{|c|c|c|}
%\hline
%Benchmark & Avg \# Test Cases per Task & Reference Solutions\\
%\hline
%APPS & ?& ?\\
%\hline
%HumanEval & 7.7& -\\
%\hline
%MBPP & 3 & 1\\
%\hline
%MathQA-Python & 3 & -\\
%\hline
%ClassEval & 33.1 & 1\\
%\hline
%CoderEval & varies by branch coverage & 1\\
%\hline
%Multipl-E & 3 to 7 & -\\
%\hline
%DS-1000 & 1.6 & 1\\
%\hline
%HumanEval+ & 774.8 & -\\
%\hline
%CONCODE & - & 1\\
%\hline
%R-benchmark & - & 1\\
%\hline
%\end{tabular}
%\end{center}
%\end{table}

\subsection{Usability and Productivity} 

There has been very little research on other quality attributes for evaluating LLMs as code generation tools. However, we are aware of the work of Miah and Zhu \cite{c15} on the usability of ChatGPT and that of Ziegler et al \cite{r12} and Xu et al \cite{r13} on productivity. 

\emph{Usability} is about how easy it is to use the tool to achieve the user's goal. In \cite{c15}, it is assumed that LLM responses are checked by the human user to see if the solution can be used for his/her programming task. It need not be correct but it must be good enough to use. This means a generated code to be checkable by the user. Thus, it should be understandable. This entails that the code should be \emph{well structured} and \emph{logically clear}. The generated code should also be easy to revise and adapt. This entails that it should be \emph{concise}, \emph{complete}, and \emph{accurate} in term of close to correct, etc. Moreover, the text explanations generated by the ML model in company with the code should provide \emph{sufficient explanation} and \emph{readable}, etc. These were the quality attributes used by Miah and Zhu in evaluation of usability\cite{c15}. These quality attributes were manually assessed on a Likert scale of 1 to 5, with 1 the poorest and 5 the best quality, by comparing with a reference solution. Accuracy, however, may require execution of the generated code and comparison of the outputs against the standard answer and the outputs from the reference solution. In addition to these quality attributes, they also measured the \emph{task completion time} and the \emph{number of attempts} that the user queries ChatGPT till a satisfactory solution is obtained. 

Another aspect of usability that Miah and Zhu studied \cite{c15} is \emph{learnability}, which refers to how easy the user can learn to use the tool. They found that users can hardly improve their skills of using ChatGPT through experiences. 

Ziegler et al. investigated various aspects of productivity, including \emph{task completion time}, \emph{product quality}, \emph{cognitive load}, \emph{enjoyment}, and \emph{learning}, in their evaluation of GitHub's Copilot \cite{r12}. They measured objective observations on the usage of Copilot and compared this with questionnaire surveys filled out by the users. They found that the acceptance rate of shown solutions is a better productivity predictor than other metrics. 

In \cite{r13}, Xu et al. studied the usability of a plug-in to Python's PyCharm IDE, which enables both code generation based on a ML model and code retrieval via a search of the internet. They conducted controlled experiments with human users to compare the impact of both of these on productivity in terms of \emph{task completion time} and quality of the result code in terms of \emph{correctness score}, which is assessed manually according to a marking rubric. They also used the lengths of initial and final l codes and the \emph{editing distances} to measure the quality of the generated/retrieved code.

\subsection{Multi-Trial vs Multi-Attempt Metrics} \label{sec:Metrics}

In \cite{c17}, Kulal et al. asked the ML model to generate 100 solutions on each coding task, and regarded the code generation as successful if at least one solution passed all test cases. This was later generalised to requiring $k>0$ solutions for each task, leading to the $pass@k$ metric, which is the probability of  generating at least one solution successfully in $k$ trials. Chen et al. \cite{c9} found that a straightforward calculation of the metric by its definition produces a high variance, however, so instead of recording whether there is a successful solution or not, they count the number $c$ of successful solutions in $k$ and use $c$ and $k$ to make an unbiased estimation of the $pass@k$ metric.  This approach has been used by most of the benchmarks reported above, including ClassEval, MBPP, MathQA-Python, CoderEval, and HumanEval+. 

The $pass@k$ metrics is applicable to the testing and evaluation processes where the ML model is tried multiple times on each test case. Thus, it is called a \emph{multi-trial metric}.

Miah and Zhu \cite{c15}, in contrast, consider the task of code generation from an LLM model, ChatGPT specifically, to be an interactive process in which the user makes a number of attempts by entering and amending their input to the LLM until a successful solution is generated, or the user gives up after a certain number $k$ of allowed attempts without obtain a satisfactory solution. They proposed a new metric $\#attempt_k$ that measures the average number of attempts and found in their experiments that satisfactory solutions from ChatGPT can be obtained with 1.6 attempts on average. 

Table \ref{tabV} provides detailed information regarding the benchmarks used in the evaluation, the metrics used and the main results. 

\begin{table}[h]
\caption{Evaluation per Benchmark}
\label{tabV}
\begin{center}
\begin{scriptsize}
\begin{tabular}{|l|l|l|l|c|}
\hline
\textbf{Benchmark} &\textbf{Correctness} &\textbf{Metric} &\textbf{Model} &\textbf{Result}\\
\hline
APPS &  pass all tests    & \%pass@100    &GPT-2 1.5B &0.68\\
           &                           &                          &GPT-Neo 2.7B &1.12\\
           &                           &                          &GPT-3 175B &0.06\\
           &  test pass rate   &    AvgTPR         &GPT-2 1.5B &7.96 \\
           &                           &                          &GPT-Neo 2.7B &10.15 \\
           &                           &                          &GPT-3 175B & 0.55  \\
\hline
HumanEval & pass all tests & \%pass@100 
%&GPT-Neo-125M &2.97 \\
%                   &                       &                    &GPT-Neo-1.3B &16.30 \\
%                   &                       &                    
                                                                   &GPT-Neo-2.7B &21.37\\
                   &                       &                    &GPT-J-6B &27.74\\
                   &                       &                    &Tabnine &7.59\\
%                   &                       &                    &Codex-12M &8.58\\
%                   &                       &                    &Codex-25M &12.89\\
%                  &                       &                    &Codex-42M &15.55\\
%                   &                       &                    &Codex-85M &22.4\\
%                   &                       &                    &Codex-300M &36.27\\
%                   &                       &                    &Codex-679M &40.95\\
%                   &                       &                    &Codex-2.5B &59.5 \\
                   &                       &                    &Codex-12B &72.31\\
\hline
MBPP  & pass all tests & \%pass@1 &Decoder only-8B  &79.0 \\
			  &                      &                    &Transformer-68B &82.8 \\
			  &                      &                    &Lang. Model-137B &83.8 \\
\hline
MathQA & pass all tests & \%pass@1 &Decoder only-8B &74.7\\
-Python &                        &                   &Transformer-68B &79.5 \\
             &                         &                   &Lang. Model-137B  &81.2 \\
\hline
ClassEval & pass all tests & \%pass@5 &  GPT-4 & 42.0 \\
                 &                       &                   &  GPT-3.5 &36.0 \\
                 &                       &                   &  WizardCoder &23.0 \\
                 &                       &                   &  StarCoder &14.0 \\
                 &                       &                   &  SantaCoder &10.0 \\
                 &                       &                   &  CodeGen &13.0 \\
                 &                       &                   &  CodeGeeX &10.0 \\
                 &                       &                   &  InCoder &8.0 \\
                 &                       &                   &  Vicuna &4.0 \\
                 &                       &                   &  ChatGLM &3.0 \\
                 &                       &                   &  PolyCoder &3.0 \\
\hline
CoderEval & pass all tests & \%pass@10 &CodeGen &23.48 \\
(Python)    &                       &                      &PanGu-Coder &27.39 \\  
                  &                       &                      &ChatGPT &30.00 \\
\hline
CoderEval &pass all tests & \%pass@10 &CodeGen &33.48 \\  
(Java)        &                       &                     &PanGu-Coder &43.04 \\  
                  &                       &                     &ChatGPT &46.09 \\
\hline
Multipl-E        &pass all tests & \%pass@1 &Codex &$\approx$ 36 \\
(HumanEval) &                      &                   &CodeGen &$\approx$ 9\\
                      &                      &                   &InCoder &$\approx$ 6 \\
\hline
Multipl-E  & pass all tests & \%pass@1 &Codex & $\approx$40\\
(MBPP)    &                      &                    &CodeGen & $\approx$14 \\  
                &                      &                    &InCoder &$\approx$ 15\\
\hline
DS-1000 & pass all tests & \%pass@1 &Codex-002 &41.25 \\  
%               &                       &                    &Codex-001 &20.2 \\  
%               &                       &                    &Codex-Cushman &18.1 \\  
               &                       &                    &CodeGen-6B &8.4 \\  
               &                       &                    &InCoder-6B &7.45 \\
\hline
HumanEval+ & pass all tests & \%pass@100 &CodeGen &$\approx$64.0 \\  
                      &                      &                        &SantaCoder &40.6 \\  
                      &                      &                        &InCoder &$\approx$29.8 \\  
                      &                      &                        &PolyCoder &13.6 \\  
                      &                      &                        &ChatGPT & 89.8\\  
                      &                      &                        &Vicuna &40.25 \\  
                      &                      &                        &StableLM-$\alpha$ &11.9 \\  
                      &                      &                        &GPT-J &25.9 \\  
                      &                      &                        &GPT-Neo &16.8 \\
\hline
ConCode & BLEU & Avg BLEU &Retrieval &20.27 \\  
                 &            &                  &Seq2Seq &23.51 \\  
                 &            &                  &Seq2Prod &21.29 \\  
                 &            &                  &User-designed &22.11 \\
\hline
R-benchmark & Satisfactory & Avg \#attempt$_k$ & ChatGPT & 1.6\\
\hline
\end{tabular}
\end{scriptsize}
\end{center}
\end{table}

\section{Research Directions} \label{sec:Directions}

In the past few years, significant progress has been made both in the development of benchmarks for code generation and in techniques for evaluation. However, there are a few problems that require further research. 

First, as shown in Table \ref{tabI}, most benchmarks were constructed from a single source, so they lack diversity and their distribution may be skewed towards certain types of questions. %For example, Stack Overflow problems may have a different distribution of the problems in textbooks and vice versa. 

Second, using tests to judge correctness of a generated solution has the advantages of being objective and automatic. However, the accuracy of the judgement depends heavily on test adequacy. This has, however, been reported in more recent benchmarks. For example, ClassEval's test suites have an average of 98.2\% branch coverage and 99.7\% statement coverage. CoderEval aims to provide full branch coverage. HumanEval+ claims to encompass all possible corner cases. A further problem of judging correctness by test results that has not been noticed by the research community is that pre-scripted test cases  can detect only errors of omission, as opposed to errors of commission, which can include malicious code. 

The overall performance metric $pass@k$, the probability of getting at least one correct solution in $k$ outputs, reflects the randomness of the LLM output. However, as Miah and Zhu pointed out, users do not normally run the LLM several times so $pass@k$ does not reflect its usability. 
%For example, the evaluation scores on $pass@k$ were low, e.g. 28.8\% on $pass@100$ for Codex, while Codex as the model underlying ChatGPT remain popular among programmers. 
The $\#attempts_k$ metric seems to be a better fit but it heavily relies on manual assessment and subjective judgement so it is difficult to apply to large scale experiments. 

The validity problems of syntax similarity metrics were studied in ML models before LLMs were introduced. It is possible that these problems still exist. Ideally, these metrics should be modified to measure usability as this is more important than correctness for LLMs \cite{r12}. 

Finally, with the exception of  the R-benchmark \cite{c15}, existing benchmarks do not support scenario-based evaluation. A single evaluation score is given and that does not tell the developer how to improve the model. The problem is how to perform scenario-based evaluation effectively and efficiently so that better feedback can be given. Including metadata in benchmarks seems to be a promising approach. However, manually assigning this metadata to coding tasks in a large scale dataset like in \cite{c15} is labour intensive, time consuming and costly. The challenge is to do it automatically.

\section{Conclusion}\label{sec:Conclusion}

Evaluation of LLMs as intelligent code generation tools is still a grave challenge. There are many open problems in the construction of benchmarks and the definition and implementation of performance metrics despite of the great efforts reported recently in the literature. Among the most important problems to be solved are the development of performance metrics that reflect ML model's usability, the validation of the metrics, the construction of benchmarks that are versatile and feasible to use,  and the techniques and tools that enable the automation of evaluation. 

\addtolength{\textheight}{-13cm}   % This command serves to balance the column lengths

%%%%%%%%%%%%%%%%%%%%%%%%%%%%%%%%%%%%%%%%%%%%%%%%%%%%%%%%%%%%%%%%%%%%%%%%%%%%%%%%

\end{document}